\documentclass{IEEEtran}
\usepackage{fullpage}
\usepackage[ruled,vlined,linesnumbered,boxruled]{algorithm2e}
\usepackage{mathrsfs}
\usepackage{graphicx}
\usepackage{amsfonts}
\usepackage{amsmath}
\usepackage{amssymb}
\usepackage{array}
\usepackage{flafter}
\usepackage{subfigure}
\usepackage{verbatim}
\usepackage[usenames]{color}
\usepackage[svgnames]{xcolor}
\usepackage{hyperref}

\usepackage{amsthm}	
\usepackage{etoolbox}

\providecommand{\citet}[1]{\citeauthor{#1}\,[\citeyear{#1}]}
\providecommand{\citep}[1]{\cite{#1}}

\newtheorem{Lem}{Lemma}

\newtheorem{assumption}{Assumption}

\newtheorem{Remark}{Remark}

\newtoggle{finalpaper}
\togglefalse{finalpaper}

\graphicspath{{./figs/}}

\allowdisplaybreaks[1]

\newcommand{\pr}[1]{\textbf{#1:} }

\newcommand{\tv}{\varpi} 

\usepackage{geometry}
\newcommand{\papermargin}{0.97in} 
\newgeometry{top=\papermargin,bottom=\papermargin,right=\papermargin,left=\papermargin}
\usepackage{enumitem}

\title{\LARGE \bf Confidence-aware Occupancy Grid Mapping: \\
	A Planning-Oriented Representation of Environment
	\iftoggle{finalpaper}
	{with Application to Stereo Vision-based Quadcopter Navigation}
	{}
}
\author {Ali-akbar Agha-mohammadi
	\thanks{Ali Agha is with the Jet Propulsion Laboratory (JPL), California Institute of Technology, Pasadena, CA 91109. 
		\{\tt\small aliakbar.aghamohammadi@jpl.nasa.gov\}}%
}

\begin{document}

\maketitle

\begin{abstract}
Occupancy grids are the most common framework when it comes to creating a map of the environment using a robot. This paper studies occupancy grids from the motion planning perspective and proposes a mapping method that provides richer data (map) for the purpose of planning and collision avoidance. Typically, in occupancy grid mapping, each cell contains a single number representing the probability of cell being occupied. This leads to conflicts in the map, and more importantly inconsistency between the map error and reported confidence values. Such inconsistencies pose challenges for the planner that relies on the generated map for planning motions. In this work, we store a richer data at each voxel including an accurate estimate of the variance of occupancy. We show that in addition to achieving maps that are often more accurate than tradition methods, the proposed filtering scheme demonstrates a much higher level of consistency between its error and its reported confidence. This allows the planner to reason about acquisition of the future sensory information. Such planning can lead to active perception maneuvers that while guiding the robot toward the goal aims at increasing the confidence in parts of the map that are relevant to accomplishing the task.
\end{abstract}

\section{Introduction} \label{subsec:intro}
Consider a Quadrotor flying in an obstacle-laden environment, tasked to reach a goal point while mapping the environment with a forward-facing stereo camera. To carry out the sense-and-avoid task, and ensure the safety of the system by avoiding collisions, the robot needs to create a representation of obstacles, referred to as the map, and incorporate it in the planning framework. This paper is concerned with the design of such a framework where there is tight integration between mapping and planning. The main advantage of such tight integration and joint design of these two blocks is that not only mapping can provide the information for the planning (or navigation) module, but also the navigation module can generate maneuvers that lead to better mapping and more accurate environment representation.


Grid-based structures are among the most common representation of the environment when dealing with the stereo cameras. Typically, each grid voxel contains a boolean information that if the cell is free or occupied by obstacles. In a bit richer format each voxel contains the probability of the cell bing occupied. Traditionally such representation is generated assuming that the robot actions are given based on some map-independent cost. However, in a joint design of planning and mapping, one objective of planning could be the accuracy of generated map.

First applications of occupancy grids in robotics date back to \cite{moravec1988sensor} and \cite{Elfes1989occupancy} and since then they have been widely used in robotics. \cite{Thrun2005}, \cite{stachniss2009_book}, and \cite{thrun2002robotic} discuss many variants of these methods. Grid-based maps have been constructed using different ranging sensors, including stereo-cameras \cite{konolige2008outdoor}, sonars \cite{yamauchi1997frontier}, laser range finders \cite{thrun1998learning}, and their fusion \cite{moravec1988sensor}. Their structure has been extended to achieve more memory efficient maps \cite{wurm2010octomap}. Further, various methods have extended grid-based mapping to store richer forms of data, including distance to obstacle surface \cite{newcombe2011kinectfusion}, reflective properties of environment \cite{howard1996generating}, and color/textureness \cite{moravec1996TR}. 

The main body of literature, however, uses occupancy grids to store binary occupancies updated by log-odds method which will be discussed in Section \ref{sec:background}. While demonstrated a high success in a variety of applications, these methods suffer three main issues, in particular when the sensory system is noisy (e.g., stereo or sonar). The first issue is that they update the occupancy of each voxel fully independent of the rest of the map. This is a very well-known problem \cite{Thrun2005} and has been shown that leads to conflicts between map and measurement data. In particular when the sensor is noisy or it has a large field of view, there is a clear coupling between voxels that fall into the field of view of the sensor. Second, these methods rely on a concept called ``inverse sensor model" (ISM), which needs to be hand-engineered for each sensor and a given environment. Third, they store a single number at each voxel to represent its occupancy. As a result, there is no consistent confidence/trust value to help the planner in deciding how reliable the estimated occupancy is.

Different researchers have studied these drawbacks and proposed methods to alleviate these issues, including \cite{pagac1996evidential}, \cite{konolige1997improved}, \cite{Paskin05}, \cite{veeck2004learning}, and \cite{thrun2003learning}. All these methods attempt to alleviate the negative effects caused by the incorrect voxel-independence assumption in mapping. In particular, \cite{thrun2003learning} proposes a grid-mapping method using forward sensor models, which takes into account all voxel dependencies and achieve maps with higher quality compared to maps resulted from ISM. However, it requires the measurement data to be collected offline and runs an expectation maximization on the full data to compute the most likely map.

\textit{Contributions and highlights:} In this paper we review traditional mapping methods and its assumptions. Accordingly, we propose a method that aims at relaxing these assumptions and generates more accurate maps, as well as more consistent filtering mechanism. The features of the proposed method and contributions can be listed as follows:
\begin{enumerate}[leftmargin=0cm,itemindent=.5cm,labelwidth=0.4cm,labelsep=0cm,align=left]
	\item The main assumption in traditional occupancy grid mapping is (partially) relaxed: we take into account the dependence between voxels in the measurement cone at every step.
	\item The ad-hoc inverse sensor model is replaced by a so-called ``sensor cause model" that is computed based on the forward sensor model in a principled manner.
	\item In addition to the most likely occupancy value for each voxel, the map contains confidence values (e.g., variance) on voxel occupancies. The confidence information is crucial for planning over grid maps. Sensor model and uncertainties are incorporated in characterizing map accuracy.
	\item The proposed method can also relax binary assumption on the occupancy level, i.e., it is capable of coping with maps where each voxel might be partially occupied by obstacles.
	\item Compared to more accurate batch methods this method does not require logging the data in an offline phase and the map can be updated online as the sensory data is received.
	\item While the main focus of this paper is on the mapping part, we discuss a planning framework where active perception/mapping is accomplished via incorporating the proposed mapping scheme into the planning, where the future evolution of the map under planner actions can be predicted accurately.
\end{enumerate}

\textit{Paper organization:} We start by the problem statement and a review of the log-odds based mapping method in the next two sections. In Section \ref{sec:sensorModel}, we discuss the sensor model we consider for the stereo camera. Section \ref{sec:mapping} describes our mapping framework. In Section \ref{sec:planning}, we explain the planning algorithm. Section \ref{sec:simulation} demonstrates the results of the proposed mapping method.

\section{Occupancy grid mapping using inverse sensor models}\label{sec:background}
Most of occupancy grid mapping methods decompose the full mapping problem to many binary estimation problems on individual voxels assuming full independence between voxels. This assumption leads to inconsistencies in the resulted map. We discuss the method and these assumptions in this section.

Let $ G = [G^{1},\cdots,G^{n}] $ be an $ n $-voxel grid overlaid on the 3D (or 2D) environment, where $ G^{i}\in\mathbb{R}^{3} $ is a 3D point representing the center of the $ i $-th voxel of the grid. Occupancy map $ m = [m^{1},\cdots,m^{n}]$ is defined as a set of values over this grid. We start with a more general definition of occupancy where $ m^{i}\in[0,1]$ denotes what percentage of voxel is occupied. $ m^{i}=1 $ when the $ i $-th voxel is fully occupied and $ m^{i}=0 $ when it is free. We overload the variable $ m^{i} $ with function $ m^{i}[x]=G^{i} $ that returns the 3D location of the $ i $-th voxel in the global coordinate frame.

The full mapping problem is defined as estimating map $ m $ based on obtained measurements and robot poses. We denote the sensor measurement at the $ k $-th time step by $ z_{k} $ and the sensor configuration at the $ k $-th time step with $ xv_{k} $. Formulating the problem in a Bayesian framework, we compress the information obtained from past measurements $ z_{0:k}=\{z_{0},\cdots,z_{k}  \} $ and $ xv_{0:k}=\{xv_{0},\cdots,xv_{k}  \} $ to create a probability distribution (belief) $ b^{m}_{k} $ on the map $ m $.
\begin{align}
b^{m}_{k}=p(m|z_{0:k},xv_{0:k})
\end{align}

However, due to challenges in storing and updating such a high-dimensional belief, grid mapping methods start from individual cells (marginal distributions).
\begin{assumption}
	\pr{Collection of marginals}
	Map pdf is represented by the collection of individual voxel pdfs (marginal pdfs), instead of the full joint pdf.
	\begin{align}
	b^{m}_{k}\equiv(b^{m^{i}}_{k})_{i=1}^{n},~~~~~
	b^{m^{i}}_{k}=p(m^{i}|z_{0:k},xv_{0:k})
	\end{align}
	where $ n $ denotes the number of voxels in the map.
	\label{assump:cell_indep}
\end{assumption}

To compute the marginal $ b^{m^{i}} $ in a recursive manner, the method starts with applying the Bayes rule.
\begin{align}
\label{eq:firstBayes}
b^{m^{i}}_{k}&=p(m^{i}|z_{0:k},xv_{0:k})\\
\nonumber&=
\frac{p(z_{k}|m^{i},z_{0:k-1},xv_{0:k})p(m^{i}|z_{0:k-1},xv_{0:k})}{p(z_{k}|z_{0:k-1},xv_{0:k})}
\end{align}

The main incorrect assumption is applied here:
\begin{assumption}
	\pr{Measurement independence}
	It is assumed that occupancy of voxels are independent given the measurement history. Mathematically:
	\begin{align}
	p(z_{k}|m^{i},z_{0:k-1},xv_{0:k})\approx p(z_{k}|m^{i},xv_{k})\label{eq:wrong_indpendece}
	\end{align}
	\label{assump:measurement_independence}
\end{assumption}

\begin{Remark}
	Note that Assumption \ref{assump:measurement_independence} would be precise if conditioning was over the whole map. In other words, 
	\begin{align}
	p(z_{k}|m,z_{0:k-1},xv_{0:k})= p(z_{k}|m,xv_{k})
	\end{align}
	is correct. But, when conditioning on a single voxel, approximation could be very off, because a single voxel $ m^{i} $ is not enough to generate the likelihood of observation $ z $. For example, there might even be a wall between $ m^{i} $ and the sensor, and clearly $ m^{i} $ alone cannot tell what range will be measured by the sensor in that case.
\end{Remark}

\begin{Remark}
	The approximation is tight for accurate sensors such as lidars, because the sensor is very accurate and the measurement likelihood function looks like a delta function. Therefore, likelihood is close to one if the range measurement is equal to the distance of $ m^{i} $ from sensor, and is zero, otherwise; Hence, the success of ISM-based methods in such settings. But, when dealing with noisy sensors such as stereo cameras or even in the absence of noise when dealing with sensors with large measurement cone (such as sonar) this assumption leads to conflicts in the map and estimation inconsistency.
\end{Remark}

\pr{Inverse sensor model}
Following Assumption \ref{assump:measurement_independence}, one can apply Bayes rule to Eq. \ref{eq:wrong_indpendece}
\begin{align}\label{eq:inverseSensorModel}
p(z_{k}|m^{i},xv_{k})=
\frac{p(m^{i}|z_{k},xv_{k})p(z_k|xv_k)}{p(m^{i}|xv_{k})}
\end{align}
which gives rise to the concept of inverse sensor model, i.e., $ p(m^{i}|z_{k},xv_{k}) $. Inverse sensor model describes the occupancy probability given a single measurement. The model cannot be derived from sensor model. However, depending on application and the utilized sensor, ad-hoc models can be hand-engineered. The reason to create this model is that it leads to an elegant mapping scheme on binary maps as follows.

Plugging \eqref{eq:wrong_indpendece} and \eqref{eq:inverseSensorModel} into \eqref{eq:firstBayes}, we get:
\begin{align}
b^{m^{i}}_{k}&=p(m^{i}|z_{0:k},xv_{0:k})\\
\nonumber
&=
\frac{p(m^{i}|z_{k},xv_{k})p(z_k|xv_k)p(m^{i}|z_{0:k-1},xv_{0:k})}{p(m^{i}|xv_{k})p(z_{k}|z_{0:k-1},xv_{0:k})}
\end{align}

Given that robot's motion does not affect the map:
\begin{align}\label{eq:blahblah}
b^{m^{i}}_{k}&=p(m^{i}|z_{0:k},xv_{0:k})\\
\nonumber
&=
\frac{p(m^{i}|z_{k},xv_{k})p(z_k|xv_k)p(m^{i}|z_{0:k-1},xv_{0:k-1})}{p(m^{i})p(z_{k}|z_{0:k-1},xv_{0:k})}
\end{align}

\begin{assumption}
	\pr{Binary occupancy}
	To complete the recursion, it further is assumed that the occupancy of voxels are binary. We denote the binary occupancy by $ o^{i} \in\{0,1\} $. Thus, $ p(o^{i}=1)=1-p(o^{i}=0) $.
	\label{assump:binary_occupancy}
\end{assumption}

According to Assumption \ref{assump:binary_occupancy}, one can define odds $ r^{i}_{k} $ of occupancy and compute it using Eq.\eqref{eq:blahblah}:
\begin{align}
\label{eq:logOdds_definition}
r^{i}_{k}&:=\frac{p(o^{i}=1|z_{0:k},xv_{0:k})}{p(o^{i}=0|z_{0:k},xv_{0:k})}\\
\nonumber
&=
\frac{p(o^{i}=1|z_{k},xv_{k})p(m^{i}=0)}{p(o^{i}=0|z_{k},xv_{k})p(m^{i}=1)}r^{i}_{k-1}
\end{align}

\begin{Remark}
	Making Assumption \ref{assump:binary_occupancy} and using odds, removes difficult-to-compute terms from the recursion in Eq. \eqref{eq:blahblah}.
\end{Remark}

\begin{figure}[h!]
	\centering
	\includegraphics[width=0.8\columnwidth]{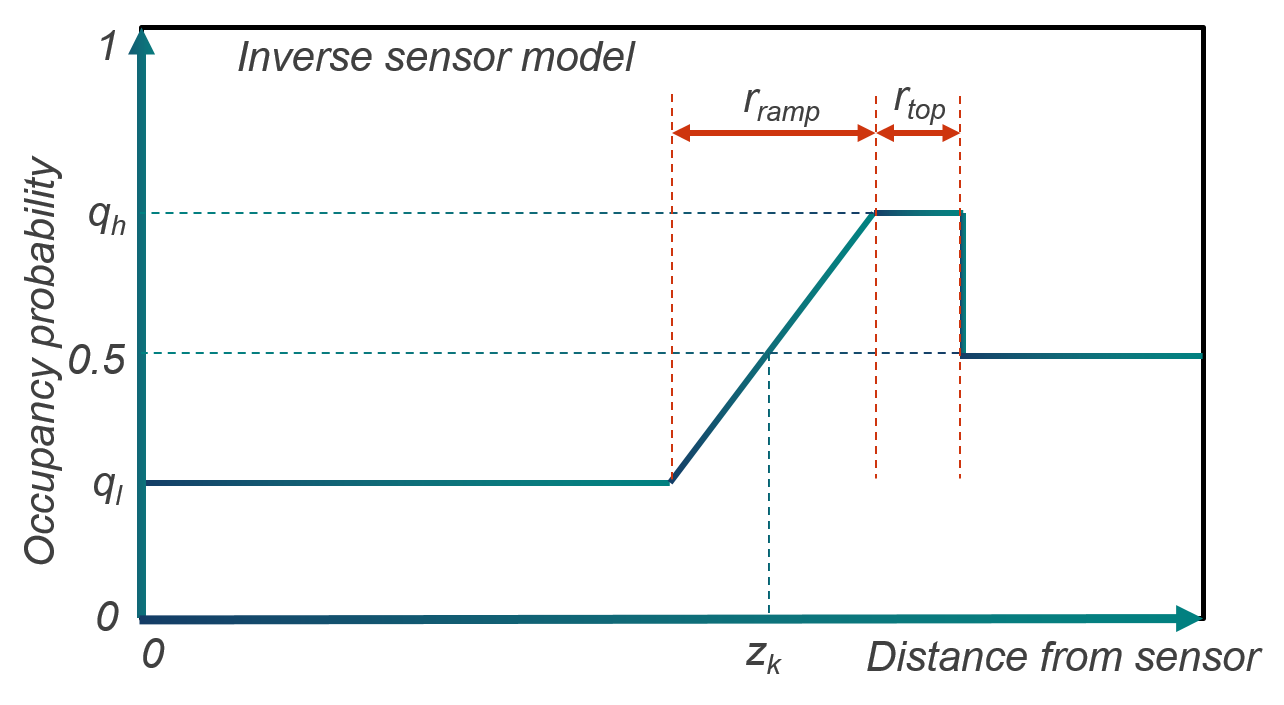}
	\caption{Typical inverse sensor model for a range sensor. It returns the occupancy probability for voxels on the measurement ray/cone based on their distance to camera.}
	\label{fig:invSensorModel}
\end{figure}

Further, denoting log-odds as $ l_{k}^{i}=\log r^{i}_{k} $, we can simplify the recursion as:
\begin{align}\label{eq:logOdds-recursion}
l^{i}_{k}&=l^{i}_{k-1}+l^{i}_{ISM}-l_{prior}
\end{align}
where, $ l_{ISM}^{i}=p(o^{i}=1|z_{k},xv_{k})p(o^{i}=0|z_{k},xv_{k})^{-1} $ is the log-odds of ISM at voxel $ i $, and $ l^{i}_{prior}=p(o^{i}=1)p(o^{i}=0)^{-1} $ is the log-odds of prior. ISM is often hand-engineered for a given sensor/environment. Fig. \ref{fig:invSensorModel} shows the typical form of ISM function.

\section{Confidence-rich Representation} \label{subsec:problem}
In this paper, we store the probability distribution of $ m^{i} $ in each voxel $ i $. Variable $ m^{i} $ in this paper can be interpreted in two ways:
\begin{enumerate}[leftmargin=0cm,itemindent=.5cm,labelwidth=0.4cm,labelsep=0cm,align=left]
	\item If the underlying true map is assumed to be a binary map, the occupancy of the $ i $-th voxel $ o^{i}\in\{0,1\} $ is distributed as Bernoulli distribution $ o^{i}\sim Bernoulli(m^{i}) $. In this case $ m^{i} $ refers to the parameter of the Bernoulli distribution. While inverse sensor-based mapping methods store $ m^{i} $ as a deterministic value, we estimate $ m^{i} $ probabilistically based on measurements and store its pdf at each voxel.
	\item The proposed method can also model continuous occupancy. In that case $ m^{i}\in[0,1] $ directly represents the occupancy level (the percentage of voxel $ i $ that is occupied by obstacles.) All below machinery applies to this case, and Assumption \ref{assump:binary_occupancy} in occupancy mapping can be relaxed. 
\end{enumerate}
However, to keep the discussion coherent, in below presentation we follow the first case: binary occupancy case $ o^{i}\in\{0,1\} $, where $ m^{i}\in[0,1] $ represents the occupancy probability, i.e., $ m^{i}_{k}=p(o^{i}=1|z_{0:k},xv_{0:k}) $.

\pr{Problem description}
Given the above-mentioned representation, we aim at estimating $ m $ based on noisy measurements by computing its posterior distribution $ b^{m}_{k}=p(m|z_{0:k},xv_{0:k}) $. Similar to ISM-mapping, we only keep marginals, i.e., $ b^{m}_{k}\equiv(b^{m^{i}}_{k}) $, for all $ i $, where $ b^{m^{i}}_{k}=p(m^{i}|z_{0:k},xv_{0:k}) $. To do so, we derive the following items:

\begin{enumerate}[leftmargin=0cm,itemindent=.5cm,labelwidth=0.4cm,labelsep=0cm,align=left]
	\item \pr{Ranging sensor model}
	Given the obstacles are described by a stochastic map, we derive a ranging sensor model, i.e., the probability of getting measurement $ z $ given a stochastic map and robot location: $ p(z_{k}| xv_{k}, b^{m}_{k}) $. This model will be utilized in map update module.
	
	\item \pr{Recursive density mapping}
	We derive a recursive mapping scheme $ \tau $ that updates the current density map based on the last measurements
	\begin{align}
	b^{m^{i}}_{k+1}=\tau^{m^{i}}(b^{m}_{k},z_{k+1},xv_{k+1})
	\end{align}
	The fundamental difference with ISM-mapping is that the evolution of the $ i $-th voxel depends on other voxels as well. Note that the input argument to $ \tau^{m^{i}} $ is the full map $ b^{m} $, not just the $ i $-th voxel map $ b^{m^{i}} $. 
	
	\item \pr{Motion planning and active perception in density maps}
	While planning is beyond the scope of this method, we briefly discuss how planning can benefit from this enriched map data, to generate actions that actively reduce uncertainty on the map and leads to safer paths.
	\begin{align}
	\pi^* = \arg\min_{\Pi} J(x_{k}, b^{m}_{k}, \pi)
	\end{align}
\end{enumerate}

Overall, this method relaxes Assumptions \ref{assump:measurement_independence} and \ref{assump:binary_occupancy} of the ISM-based mapping.

\section{Range-sensor Modeling} \label{sec:sensorModel}
In this section, we model a range sensor when the environment representation is a stochastic map. We focus on passive ranging sensors like stereo cameras, but the discussion can easily be extended to active sensors too. 

\pr{Ranging pixel}
Let us consider an array of ranging sensors (e.g., disparity pixels). We denote the camera center by $ x $, the 3D location of the $ i $-th pixel by $ v $, and the ray emanating from $ x $ and passing through $ v $ by $ xv = (x,v) $. Let $ r $ denote the distance between the camera center and the closest obstacle to the camera along ray $ xv $. In stereo camera range $ r $ is related to the measured disparity $ z $ as:
\begin{align}
z = r^{-1}fd_b
\end{align}
where, $ f $ is camera's focal length and $ d_b $ is the baseline between two cameras on the stereo rig. In the following, we focus on a single pixel $ v $ and derive the forward sensor model $ p(z|xv,b^{m}) $.


\pr{Pixel cone}
Consider the field of view of pixel $ v $. Precisely speaking, it is a narrow 3D cone with apex at $ x $ and boundaries defined by pixel $ v $. Also, for simplicity one can consider just a ray $ xv $ going through camera center $ x $ and the center of pixel $ v $. Pixel cone $ \mathtt{Cone}(xv) $ refers to the set of voxels in map $ m $ that fall into this cone (or lie on ray $ xv $). We denote this set by $ \mathbb{C}=\mathtt{Cone}(xv) $.

\begin{figure}[h]
	\centering
	\includegraphics[width=0.45\textwidth]{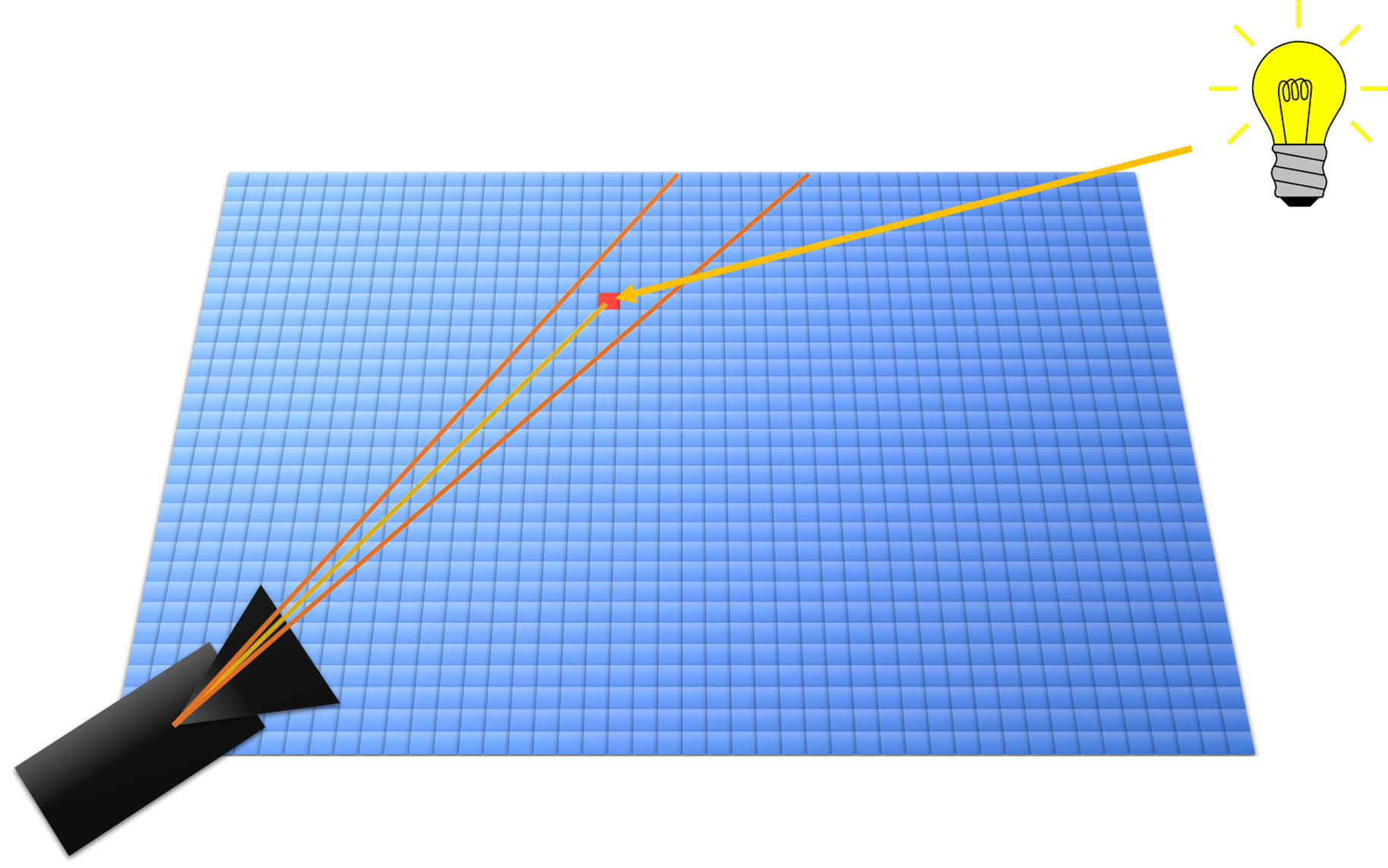}
	\caption{Cone formed by two red lines depicts the field of view of pixel $ v $. The disparity measurement on pixel $ v $ can be caused by light bouncing off any of voxels in the pixel cone and reaching the image plane. In this figure, the measurement is created by light bouncing off the ``red" voxel.}
	\label{fig:grid}
\end{figure}

\pr{Cause variables}
The disparity measurement on pixel $ v $ could be the result of light bouncing off any of voxels in the cone $ \mathbb{C}=\mathtt{Cone}(xv) $ (see Fig. \ref{fig:grid}). Therefore any of these voxels is a potential cause for a given measurement. In the case that the environment map is perfectly known, one can pinpoint the exact cause by finding the closest obstacle to the camera center. But, when the knowledge about the environment is partial and probabilistic, the best one can deduce about causes is a probability distribution over all possible causes in the pixel cone $ \mathbb{C}=\mathtt{Cone}(xv) $. These causes will play an important role (as hidden variables) in deriving sensor model for stochastic maps.

\pr{Local vs global indices}
For a given ray $ xv $, we order the voxels along the ray from the closest to the camera to the farthest from the camera. Let $ i^{l}\in\{1,\cdots,\|\mathbb{C}\|\} $ denote the local index of a voxel on ray $ xv $. Function $ i^{g} = g(i^{l},xv) $ returns the global index $ i^{g} $ of this voxel in the map.

\pr{Cause probability}
To derive the full sensor model, we need to reason about which voxel was the cause for a given measurement. For a voxel $ c\in\mathbb{C}(xv) $ to be the cause, two events need to happen: \textit{(i)} $ B^c $, which indicates the event of light bouncing off voxel $ c $ and \textit{(ii)} $ R^c $, which indicates the event of light reaching the camera from voxel $ c $.
\begin{align}
p(c | b^{m}) \!\!=\!\! \Pr(B^{c}\!,R^{c}| b^{m}) 
\!\!=\!\!
\Pr(R^{c}|B^{c}\!, b^{m})\!\Pr(B^{c}| b^{m})
\end{align}

\pr{Bouncing probability}
To compute the bouncing probability, we rely on the fact that $ \Pr(B^{c}|m^{c}) = m^{c} $ (by the definition). 
Accordingly:
\begin{align}
\nonumber\Pr(B^{c}| b^{m}) &= \int_{0}^{1}\Pr(B^{c}|m^{c},b^{m})
p(m^{c}| b^{m})dm^{c}\\
&= \int_{0}^{1}m^{c}
b^{m^{c}}dm^{c}
=\mathbb{E}{m^{c}}=\widehat{m}^{c}
\end{align}
Note that $ \Pr(B^{c}|m^{c},b^{m}) = \Pr(B^{c}|m^{c}) $.

\pr{Reaching probability}
For the ray emanating from voxel $ c $ to reach the image plane, it has to go through all voxels on ray $ xv $ between $ c $ and sensor. Let $ c^{l} $ denotes local index of voxel $ c $ along the ray $ xv $, i.e., $ c^{l}=g^{-1}(c,xv) $, then we have:
\begin{align}
&\Pr(R^{c}| B^{c}, b^{m}) \\
\nonumber
&
\!\!
= 
\!\!(1 \!\!-\!\! \Pr(B^{g(c^{l}- 1,xv)}| b^{m}))
\!
\Pr(R^{g(c^{l}- 1,xv)}| B^{g(c^{l}- 1,xv)}\!,\! b^{m})
\\
\nonumber
&= \prod_{l = 1}^{c^{l}-1}
(1 - \Pr(B^{g(l,xv)}| b^{m}))
= \prod_{l = 1}^{c^{l}-1}
(1 - \widehat{m}^{g(l,xv)})
\end{align}

\pr{Sensor model with known cause}
Assuming the cause voxel for measurement $ z $ is known, the forward sensor is typically modeled as:
\begin{align}
z = h(xv,c,n_{z}) = \|G^{c}-x\|^{-1}fd_{b}+n_{z},
\end{align}
where, $ n_{z}\sim\mathcal{N}(0,R) $ denotes the observation noise, drawn from a zero-mean Gaussian with variance $ R $. We can alternatively describe the observation model in terms of pdfs as follows:
\begin{align}
p(z | xv,c) = \mathcal{N}(\|G^{c}-x\|^{-1}fd_{b}, R)
\end{align}

\pr{Senosr model with stochastic maps}
Sensor model given a stochastic map can be computed by incorporating hidden cause variables into the formulation:
\begin{align}
&p(z | xv;b^{m}) = 
\sum_{c\in\mathbb{C}(xv)}p(z|xv,c;b^{m})\Pr(c|b^{m})\\
\nonumber
&=
\sum_{c\in\mathbb{C}(xv)}
\mathcal{N}(\|G^{c}-x\|^{-1}fd_{b}, R)
\widehat{m}^{c}
\prod_{l = 1}^{c^{l}-1}
(1 - \widehat{m}^{g(l,xv)})
\end{align}

\section{Confidence-Augmented Grid Map} \label{sec:mapping}
In this section, we derive the mapping algorithm that can reason not only about the occupancy at each cell, but also about the confidence level of this value. As a result, it enables efficient prediction of the map that can be embedded in planning and resulted in safer plans.

We start by a lemma that will be used in derivations. See Appendix \ref{app:proofLemma} for proof.
\begin{Lem}\label{lem:cause-meas}
	Given the cause, the value of the corresponding measurement is irrelevant.
	\begin{align}
	\nonumber
	p(m^i|c_k,z_{0:k},xv_{0:k})
	=
	p(m^i|c_{k},z_{0:k-1},xv_{0:k})
	\end{align}
\end{Lem}

To compute the belief of the $ i $-th voxel, denoted by $ b^{m^{i}}_{k} = p(m^i|z_{0:k},xv_{0:k}) $, we bring the cause variables into formulation.
\begin{align}
\label{eq:BayesianUpdate}
&b^{m^{i}}_{k} = p(m^i|z_{0:k},xv_{0:k})\\
\nonumber
&=
\!\!\!\!\!
\sum_{c_{k}\in\mathbb{C}(xv)}
\!\!\!\!\!
p(m^i|c_k,z_{0:k},xv_{0:k})\Pr(c_k|z_{0:k},xv_{0:k})
\\
\nonumber
&=
\!\!\!\!\!
\sum_{c_{k}\in\mathbb{C}(xv)}
\!\!\!\!\!
p(m^i|c_k,z_{0:k-1},xv_{0:k})\Pr(c_k|z_{0:k},xv_{0:k})
\\
\nonumber
&=
\!\!\!\!\!
\sum_{c_{k}\in\mathbb{C}(xv)}
\!\!\!\!\!
\frac{
	\Pr(c_k|m^i,z_{0:k-1},xv_{0:k})
}
{\Pr(c_k|z_{0:k-1},xv_{0:k})}
\Pr(c_k|z_{0:k},xv_{0:k})
b^{m^{i}}_{k-1}
\end{align}

It can be shown that $ b_{k-1}^{m} $ is sufficient statistics \cite{kay1993fundamentals} for the data $ (z_{0:k-1},xv_{0:k-1}) $ in above terms. Thus, we can re-write \eqref{eq:BayesianUpdate} as:
\begin{align}
b^{m^{i}}_{k}
\!\!
=
\!\!\!\!\!\!\!
\sum_{c_{k}\in\mathbb{C}(xv)}
\!\!\!\!\!\!
\frac
{\Pr(c_k|m^i,b^{m}_{k-1},xv_{k})}
{\Pr(c_k|b^{m}_{k-1},xv_{k})}
\!
\Pr(c_k|b^m_{k-1},z_{k},xv_{k})
b^{m^{i}}_{k-1}
\label{eq:recur-with-ratio}
\end{align}

In the following, we make the assumption that the map pdf is sufficient for computing the bouncing probability from voxel $ c $ (i.e., one can ignore voxel $ i $ given the rest of the map.)
Mathematically, for $ c_{k}\neq i $, we assume:
\begin{align}
\nonumber
&\Pr(B^{c_{k}}| m^{i}, b^{m}_{k-1},xv_{k})
\!\approxeq\! \Pr(B^{c_{k}}| b^{m}_{k-1},xv_{k})
\!=\! \widehat{m}^{c_{k}}
\end{align}


Note that we still preserve a strong dependence between voxels via the reaching probability. To see this clearly, let's expand the numerator $ p(c_{k} | m^{i}, b^{m}_{k-1},xv_{k}) $ in \eqref{eq:recur-with-ratio} as (we drop $ xv $ to unclutter the equations):
\begin{align}
&p(c_{k} | m^{i}, b^{m}_{k-1},xv_{k}) 
\!\!=\!\!
\Pr(B^{c_{k}},R^{c_{k}}| m^{i}, b^{m}_{k-1},xv_{k}) 
\\
\nonumber
&= \Pr(B^{c_{k}}| m^{i}, b^{m}_{k-1},xv_{k})\Pr(R^{c_{k}}|B^{c_{k}}, m^{i}, b^{m}_{k-1},xv_{k})\\
\nonumber
&= \begin{cases}
\widehat{m}^{c_{k}}
\prod_{l = 1}^{c^{l}_{k}-1}
(1 - \widehat{m}^{g(l,xv)})
& \text{if       } c^{l}_k < i^{l}\\
m^{i}
\prod_{l = 1}^{c^{l}_{k}-1}
(1 - \widehat{m}^{g(l,xv)})
& \text{if       } c^{l}_k = i^{l}\\
\parbox[t]{0.68\columnwidth}
{
	$
	\widehat{m}^{c_{k}}
	\left(\prod_{l = 1}^{i^{l}-1}
	(1 - \widehat{m}^{g(l,xv)})\right)
	\\
	~~~~\times(1-m^{i})
	\left(\prod_{l = i^{l}+1}^{c^{l}_{k}-1}
	(1 - \widehat{m}^{g(l,xv)})\right)
	$
}
& \text{if       } c^{l}_k > i^{l}
\\
\end{cases}
\end{align}
The denominator is $ p(c_{k} | b^{m}_{k-1},xv_{k}) = \widehat{m}^{c_{k}}
\prod_{l = 1}^{c^{l}_{k}-1}
(1 - \widehat{m}^{g(l,xv)})
$ for all $ c_{k}\in\mathbb{C}(xv) $. In these equations, $ c^{l}_{k}=g^{-1}(c_{k},xv_{k}) $ and $ i^{l}=g^{-1}(i,xv_{k}) $ are the corresponding indices of $ c_{k} $ and $ i $ in the local frame.


Therefore, the ratio in \eqref{eq:recur-with-ratio} is simplified to:
\begin{align}
\nonumber
\frac
{\Pr(c_k|m^i,b^{m}_{k-1},xv_{k})}
{\Pr(c_k|b^{m}_{k-1},xv_{k})}
&=\begin{cases}
1
& \text{if       } c^{l}_k < i^{l}\\
m^{i}(\widehat{m}^{i})^{-1}
& \text{if       } c^{l}_k = i^{l}\\
(1-m^{i})(1-\widehat{m}^{i})^{-1}
& \text{if       } c^{l}_k > i^{l}\\
\end{cases}
\end{align}

Plugging the ratio back into the \eqref{eq:recur-with-ratio}, and collecting linear and constant terms, we can show that:
\begin{align}
\nonumber
&p(m^i|z_{0:k},xv_{0:k})\\
&~~~~~~=
(\alpha^{i} m^{i}+\beta^{i})p(m^i|z_{0:k-1},xv_{0:k-1})
\label{eq:final-mapping-alphaBeta}
\end{align}
where
\begin{align}
\nonumber
\alpha^{i} &=
\sum_{c^{l}_{k}=1}^{i^{l}-1}
\Pr(c_k|b^m_{k-1},z_{k},xv_{k})
\\
&+
(1-\widehat{m}^{i})^{-1}
\sum_{c^{l}_{k}=i^{l}+1}^{|\mathbb{C}(xv)|}
\Pr(c_k|b^m_{k-1},z_{k},xv_{k})
\\
\nonumber
\beta^{i} &=
(\widehat{m}^{i})^{-1}
\Pr(c_k|b^m_{k-1},z_{k},xv_{k})
\\
&-
(1-\widehat{m}^{i})^{-1}
\sum_{c^{l}_{k}=i^{l}+1}^{|\mathbb{C}(xv)|}
\Pr(c_k|b^m_{k-1},z_{k},xv_{k})
\end{align}

In a more compact form, we can rewrite Eq. \eqref{eq:final-mapping-alphaBeta} as:
\begin{align}
b_{k+1}^{m^{i}}=\tau^{i}(b_{k}^{m},z_{k+1},xv_{k+1}).
\end{align}

\pr{Sensor cause model}
The proposed machinery gives rise to the term $ \Pr(c_k|z_{0:k},xv_{0:k})
=\Pr(c_k|b^m_{k-1},z_k,xv_{k}) $, which is referred to as ``Sensor Cause Model (SCM)" in this paper. As opposed to the inverse sensor model in traditional mapping that needs to be hand-engineered, the SCM can be derived from the forward sensor model in a principled way as follows.
\begin{align}
&\Pr(c_k|z_{0:k},xv_{0:k})
=\Pr(c_k|b^m_{k-1},z_k,xv_{k})
\\
\nonumber
&=\frac{p(z_k|c_k,xv_k)\Pr(c_k|b^m_{k-1},xv_{k})}
{p(z_k|b^m_{k-1},xv_k)}\\
\nonumber
&=\eta' p(z_k|c_k,xv_k)\Pr(c_k|b^m_{k-1},xv_{k})\\
\nonumber
&=\eta' p(z_k|c_k,xv_k)\widehat{m}^{c_k}_{k-1}
\!\!\prod_{j=1}^{c^l_k-1}\!\!(1-\widehat{m}^{g(j,xv)}_{k-1})
, \forall c_k\in\mathbb{C}(xv_k)
\end{align}
where $ \eta' $ is the normalization constant.

\subsection{Confidence in Map} \label{subsec:confidenceInMap}
A crucial feature of the proposed method is that in addition to most likely map, it provides the uncertainty associated with the returned value. In doing so, it incorporates the full forward sensor model into the mapping process. In other words, it can distinguish between two voxels, where both reported as almost free (e.g., $ \hat{m}^{1}=\hat{m}^{2}=0.1 $), but one with high confidence and the other one with low confidence (e.g., $ \sigma^{m^{1}}=0.01 $ and $ \sigma^{m^{2}}=0.2 $). This confidence level is a crucial piece of information for the planner. Obviously the planner either has to avoid $ m^{2} $ since the robot is not sure if $ m^{2} $ is actually risk free (due to high variance), or the planner needs to take active perceptual actions and take another measurement from $ m^{2} $ before taking an action.

In the ISM-based method only one number is stored in the map, namely the parameter of the Bernoulli distribution. One might try to utilize the variance of the Bernoulli distribution to infer about the confidence in an ISM-based map, but due to the incorrect assumptions made in the mapping process and also since the Bernoulli distribution is a single parameter distribution (mean and variance are dependent), the computed variance is not a reliable confidence source.

It is very important to note that generally a planner can cope with large errors ``if" there is a high variance associated with it. But, if the error is high, and at the same time, filter is confident about its wrong estimate, planning would be very challenging, and prone to failure. To quantify the inconsistency between the error and reported variances in the results section, we utilize below measure:
\begin{align}
I_c = \sum_{c}ramp(|e_{c}|-2\sigma_{c})
\label{eq:inconsistencyMeasure}
\end{align}
where, $ e_c $ and $ \sigma_{c} $, respectively, denote the estimation error and variance of voxel $ c $. The ramp function $ ramp(x):=max(0,x) $ ensures that only inconsistent voxels (with respect to $ 2\sigma $) contribute to the summation. Accordingly, $ I_c$ indicates how much of the error signal is out of bound (i.e., how unreliable the estimate is) over the whole map. We will compute this measure for different maps in the Section \ref{sec:simulation}.


\iftoggle{finalpaper}
{
	
	\SetKwFunction{AnytimeHierarchicalPlanner}{AnytimeHierarchicalPlanner}
	\SetKwFunction{AddEdge}{AddEdge}
	\SetKwFunction{RefineEdge}{RefineEdge}
	\SetKwFunction{InitializeEdge}{InitializeEdge}
	\SetKwFunction{AddNode}{AddNode}
	\SetKwFunction{AddNodeSet}{AddNodeSet}
	\SetKwFunction{AddNonStationaryNode}{AddNonStationaryNode}
	\SetKwFunction{AddStationaryNode}{AddStationaryNode}
	\SetKwFunction{AnytimeConstruction}{AnytimeConstruction}
	\SetKwFunction{FIRM}{FIRM}
	
	\begin{algorithm}\label{alg:Addnode}
		\caption{Adding a Node}
		\textbf{Procedure}  : $ G^{new}\leftarrow$\AddNode$\!\!(\nu,G) $\\
		\textbf{input} : New node $ \nu $, Old graph $ G=(\mathbb{V},\mathbb{E}) $\\ 
		\textbf{output}  :  Updated graph $ G^{new} $\\ 
		{ 
			$ \check{x}\leftarrow center(\nu) $;\\ 
			Retrieve the edges into whose tube $ \nu $ falls; i.e., $ A_{\nu} = \{\zeta\in\mathbb{E}|\check{x}\in \zeta[\tv]\} $;\\ 
			\ForEach{$ \zeta\in A_{\nu} $} 
			{ 
				Add $ \nu $ to the termination set of the edge; i.e. $ \Psi^{new}\leftarrow\zeta[\Psi]\cup\{\nu\} $;\\ 
				$(\zeta,C)\leftarrow $ \RefineEdge $ (\zeta,\Psi^{new}) $;\\ $ C^{g}(\nu,\zeta)\leftarrow C $;\\ 
			} 
			Find neighboring stationary nodes of node $ \nu $; i.e., $ \mathbb{V}_{neighb} = \{\nu'\in\mathbb{V}^{s}| d(\nu,\nu')<r \} $;\\ 
			\ForEach{$ \nu'\in \mathbb{V}_{neighb} $} {
				$(\zeta,C)\leftarrow $ \AddEdge $ (\nu,\nu') $;\\ $ C^{g}(\nu,\zeta)\leftarrow C $;\\ 
				Add the new edge to the list of edges; i.e., $ \mathbb{E}\leftarrow\mathbb{E}\cup\{\zeta\} $;\\ 
			}
			\Return $ G^{new}\leftarrow(\mathbb{V},\mathbb{E}) $;
		} 
	\end{algorithm}
}{}

\section{Planning with Confidence-aware maps} \label{sec:planning}
In this section, we describe the planning method that utilizes the confidence-rich representation proposed in the previous section. 

The objective in planning is to get to the goal point, while avoiding obstacles (e.g., minimizing collision probability). To accomplish this, the planner needs to reason about the acquisition of future perceptual knowledge and incorporate this knowledge in planning. An important feature of the confidence-right map is that it enables efficient prediction of the map evolution and map uncertainty. 

\pr{Future observations}
However, reasoning about future costs, one needs to first reason about future observations. The precise way of incorporating unknown future observations is to treat them as random variables and compute their future pdf. But, a common practice in belief space planning literature is to use the most likely future observations as the representative of the future observations to reason about the evolution of belief. Let us denote the most likely observation at the $ n $-th step by:
\begin{align}
z^{ml}_{n}=\arg\max_{z} p(z|b_{n}^{m},xv_{n})
\end{align}

\pr{Future map beliefs}
Accordingly, we can compute most likely future map beliefs:
\begin{align}
b_{n+1}^{m^{i},ml}=\tau^{i}(b_{n}^{m,ml},z^{ml}_{n+1},xv_{n+1}),~~~~ n\geq k
\label{eq:ml-belief-recursion}
\end{align}
where, $ b^{m^{i},ml}_{k}=b^{m^{i}}_{k} $.

\pr{Path cost}
To assign a cost $ J(x_k,b_{k}^{m},path) $ to a given path $path = (xv_{k},u_{k},xv_{k+1},u_{k+1},\cdots,xv_{N})$ starting from $ xv_{k} $, when map looks like $ b^{m}_{k} $, one needs to predict the map belief along the path via \eqref{eq:ml-belief-recursion}. Assuming an additive cost we can get the path cost by adding up one-step costs:
\begin{align}
J(x_k,b_{k}^{m},path) = \sum_{n=k}^{N}c(b_{n}^{m,ml},xv_{n},u_{n})
\end{align}
where the cost in belief space is induced by an underlying cost in the state space, i.e.,
\begin{align}
\nonumber
&c(b_{n}^{m,ml},xv_{n},u_{n}) 
\\
&= \int c(m_n,xv_n,u_n;b^{m,ml}_{n})
b_{n}^{m,ml}(m_n)dm_n
\end{align}

\pr{One-step cost}
The underlying one-step cost $ c(m_n,xv_n,u_n;b^{m,ml}_{n}) $ depends on the application in hand. For safe navigation with grid maps, we use the following cost function:
\begin{align}
c(m,xv,u;b^m) = m^{j} + (m^{j}-\widehat{m}^{j})^2
\end{align}
where, $ j $ is the index of the cell, the robot is at. In other words, $ x\in m^{j} $.

As a result the cost in belief space will be:
\begin{align}
\nonumber
&c(b_{n}^{m},xv_{n},u_{n}) = \widehat{m}_{n}^{j}+\sigma(m^{j}_{n}) 
\\
\nonumber
&= \mathbb{E}[m^{j}_{n}|z_{k+1:n},xv_{k+1:n},b^{m}_k]
\\
&+ \mathtt{Var}[m^{j}_{n}|z_{k+1:n},xv_{k+1:n},b^{m}_k]
\end{align}
above observations are "future observation".

\iftoggle{finalpaper}
{
\pr{Cost interpretation}
The interpretation of the cost function in ?? is a key factor to understand the behavior of robot under this method.

future variance would be low...

i.e., when the robot gets there, it will not encounter surprises...

penalize surprises...

increase the confidence in map...

encourage active perception behavior that leads to a higher confidence in future map...

different from active mapping in the sense that this is task dependent...

robot will not be encouraged to explore the map for the sake of mapping... it will only explore the map as long as it is needed to accomplish the task...
}{}

\pr{Path planning}
To generate the plan we use the RRT method \cite{Lavalle01-RRT} to create a set of candidate trajectories $ \Pi = \{path^{i}\} $. For each trajectory, we compute the cost $ c(path^{i}) $ and pick the path with minimum cost.
\begin{align}
path^* = \arg\min_{\Pi} J(x_{k}, b^{m}_{k}, path)
\end{align}

\iftoggle{finalpaper}
{
	
\SetKwFunction{AnytimeHierarchicalPlanner}{AnytimeHierarchicalPlanner}
\SetKwFunction{AddEdge}{AddEdge}
\SetKwFunction{RefineEdge}{RefineEdge}
\SetKwFunction{InitializeEdge}{InitializeEdge}
\SetKwFunction{AddNode}{AddNode}
\SetKwFunction{AddNodeSet}{AddNodeSet}
\SetKwFunction{AddNonStationaryNode}{AddNonStationaryNode}
\SetKwFunction{AddStationaryNode}{AddStationaryNode}
\SetKwFunction{AnytimeConstruction}{AnytimeConstruction}
\SetKwFunction{FIRM}{FIRM}

\begin{algorithm}\label{alg:Addnode}
	\caption{Adding a Node}
	\textbf{Procedure}  : $ G^{new}\leftarrow$\AddNode$\!\!(\nu,G) $\\
	\textbf{input} : New node $ \nu $, Old graph $ G=(\mathbb{V},\mathbb{E}) $\\ 
	\textbf{output}  :  Updated graph $ G^{new} $\\ 
	{ 
		$ \check{x}\leftarrow center(\nu) $;\\ 
		Retrieve the edges into whose tube $ \nu $ falls; i.e., $ A_{\nu} = \{\zeta\in\mathbb{E}|\check{x}\in \zeta[\tv]\} $;\\ 
		\ForEach{$ \zeta\in A_{\nu} $} 
		{ 
			Add $ \nu $ to the termination set of the edge; i.e. $ \Psi^{new}\leftarrow\zeta[\Psi]\cup\{\nu\} $;\\ 
			$(\zeta,C)\leftarrow $ \RefineEdge $ (\zeta,\Psi^{new}) $;\\ $ C^{g}(\nu,\zeta)\leftarrow C $;\\ 
			} 
		Find neighboring stationary nodes of node $ \nu $; i.e., $ \mathbb{V}_{neighb} = \{\nu'\in\mathbb{V}^{s}| d(\nu,\nu')<r \} $;\\ 
		\ForEach{$ \nu'\in \mathbb{V}_{neighb} $} {
			$(\zeta,C)\leftarrow $ \AddEdge $ (\nu,\nu') $;\\ $ C^{g}(\nu,\zeta)\leftarrow C $;\\ 
			Add the new edge to the list of edges; i.e., $ \mathbb{E}\leftarrow\mathbb{E}\cup\{\zeta\} $;\\ 
			}
		\Return $ G^{new}\leftarrow(\mathbb{V},\mathbb{E}) $;
	} 
\end{algorithm}
}{}

\section{Results: Proposed Mapping Method} \label{sec:simulation}
In this section, we demonstrate the performance the proposed mapping method and compare it with the commonly used log-odds based grid mapping. We start by studying the mapping error and then we discuss the consistency the estimation process in both methods.

Figure \ref{fig:groundTruth_map} shows an example ground truth map. Each voxel is assumed to be a square with 10cm side length. The environment size is 2m-by-2m, consisting of 400 voxels. Each voxel is either fully occupied (shown in black) or empty (white). We randomly populate the voxels by 0 and 1's, except the voxels in the vicinity origin, which are set to be free in this example, to better test the mapping method. The robot's $ (x,y,\theta) $ position has been set to $ (0,0,\pi/2) $. The robot orientation is changing with a fixed angular velocity of 15 degrees per second. We run simulations for 50 seconds, almost equivalent to two full turns.

For the sensing system, we have simulated a simple stereo camera in 2D. The range of the sensor is 1 meter, with a field of view of 28 degrees. There are 15 pixels along the simulated image plane. 
Measurement frequency is 10Hz. The measurement noise is assumed to be a zero-mean Gaussian with variance $0.04$.

\newcommand{\ScenarioFigSize}{1.42in}
\begin{figure*}[ht!]
	\centering
	\subfigure{\includegraphics[width=\ScenarioFigSize]{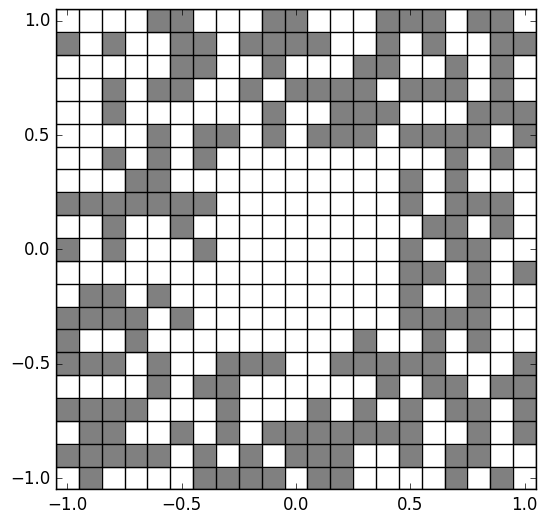}\label{fig:groundTruth_map}}
	\subfigure{\includegraphics[width=\ScenarioFigSize]{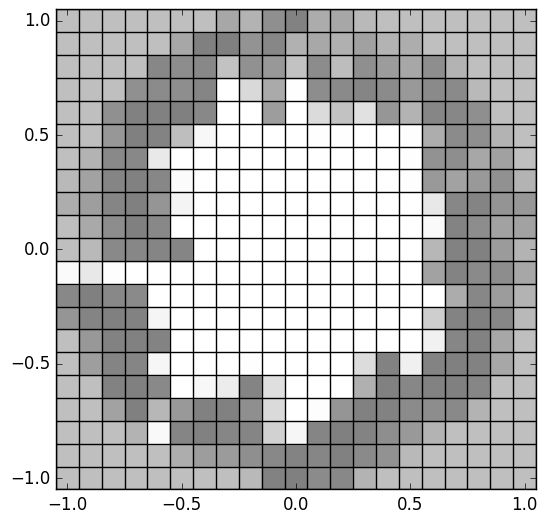}\label{fig:map_logOdd}}
	\subfigure{\includegraphics[width=\ScenarioFigSize]{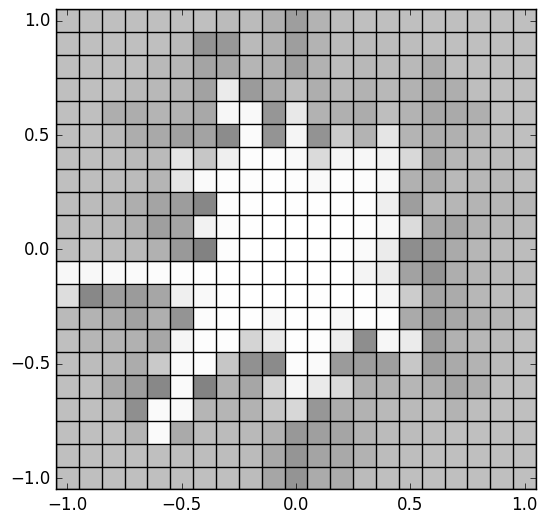}\label{fig:map_hybrid}}
	\subfigure{\includegraphics[width=1.8in]{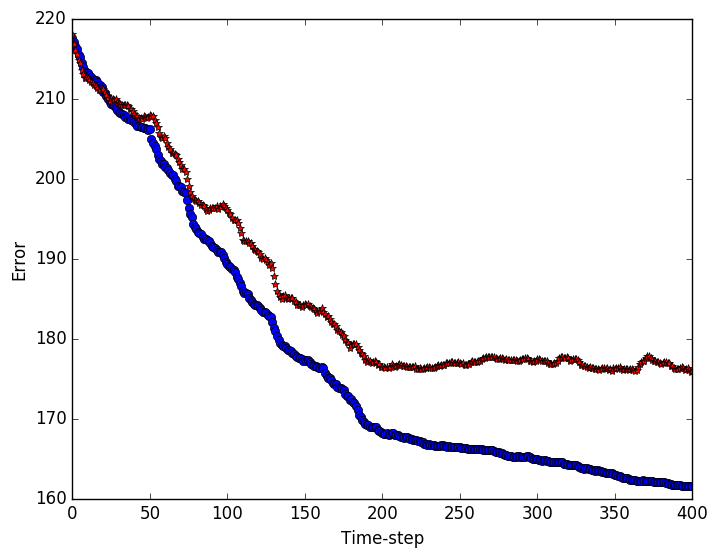}\label{fig:error_full_map}}
	\caption{(a) Ground truth map (b) Map resulted from log-odds method (c) Map resulted from the proposed method (d) Full map error evolution over time: comparison between log-odds (red) and the proposed method (blue)}
	\label{fig:example-map}
\end{figure*}

For the ISM-based mapping, we use a typical inverse sensor model as seen in Fig. \ref{fig:invSensorModel}, with parameters $ r_{ramp} = 0.1 $, $ r_{top}=0.1 $, $ q_{l}=0.45 $, and $ q_{h}=0.55 $. The map resulted from the ISM-based mapping and from the proposed method are shown in Fig. \ref{fig:map_logOdd} and \ref{fig:map_hybrid}, respectively. Note that while the inverse sensor model needs to be hand-tuned for the log-odds-based mapping, in the proposed mapping methods, there are no parameters for tuning.

To quantify the difference between the maps resulted from the log-odds and the proposed method, we compute the error between the mean of estimated occupancy and the ground truth map, in both ISM and the proposed mapping method. Then we sum the absolute value of the error over all voxels in the map as an indicator of map quality. Fig. \ref{fig:error_full_map} depicts the evolution of this value over time. As it can be seen from this figure, the proposed method shows less error than the log-odds method, and the difference is growing as more observations are obtained.

\begin{figure}[h!]
	\centering
	\includegraphics[width=0.8\columnwidth]{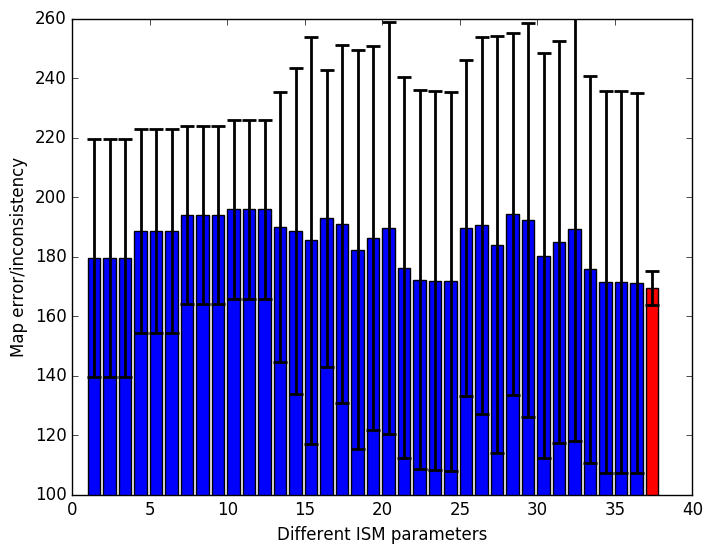}
	\caption{Error and inconsistency using different ISM parameters (blue) and the proposed method (red)}
	\label{fig:barChart}
\end{figure}

Since the inverse sensor model is typically hand-engineered for a given sensor and a given environment, we sweep over a set of inverse sensor models to compare the performance with the proposed method. Following the generic form of IMS in Fig. \ref{fig:invSensorModel}, we create 36 IMS models by \textit{(i)} setting the $ q_{h}-0.5=0.5-q_{l} $ to three values $ (0.05,0.2,0.4) $, \textit{(ii)} setting $ r_{ramp} $ to four values $ (0.05, 0.1, 0.3, 1) $, and \textit{(iii)} setting $ r_{top} $ to three values $ (0.05, 0.1, 0.3) $. Note that the voxel size is $ 0.1 $ meters. The results are shown in Fig. \ref{fig:barChart}. The height of blue bars depict the error corresponding to the 36 different ISM parameters and the red bar (right-most bar) shows the map error for the proposed method. The intervals shown in black lines depict the inconsistency measure (Eq. \ref{eq:inconsistencyMeasure}) and indicate what portion of error is a ``bad error" (inconsistent with the reported variance).

\begin{figure}[h!]
	\centering
	\includegraphics[width=0.8\columnwidth]{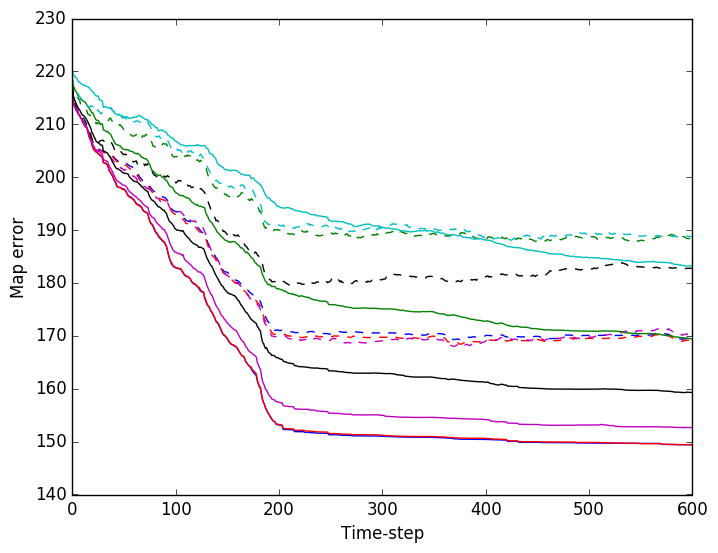}
	\caption{Comparison between ISM-based mapping (dashed lines) and the proposed method (solid lines) under different noise intensities.}
	\label{fig:stdVariation}
\end{figure}

We further study the sensitivity of the method to different sensor noises. Fig. \ref{fig:stdVariation} shows the error evolution over time for six different noise intensities/variances $ (0.1, 0.04, 0.01, 0.0025, 0.0004, 0.0001) $. Dashed lines correspond to ISM-based mapping, and solid lines correspond to the proposed method. Different noise intensities are drawn with different colors. For the same noise intensity (color) the proposed method shows less error compared to ISM.

\begin{figure}[h!]
	\centering
	\includegraphics[width=0.8\columnwidth]{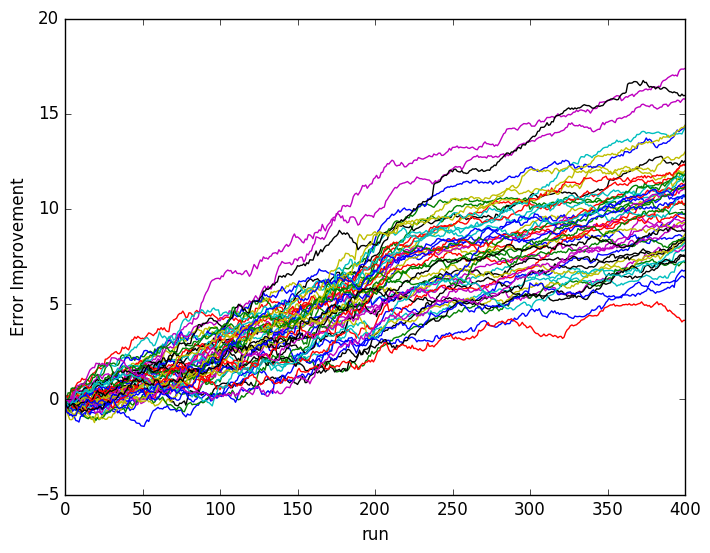}
	\caption{Performance improvement on 50 different maps.}
	\label{fig:performance}
\end{figure}

To study the improvement independent of map, we randomly create 50 environments and run the robot and mapping algorithm in these environments. Fig. \ref{fig:performance} shows the result of this Monte Carlo evaluation, which depicts in all 50 maps there is consistent improvement in the map error.

While reducing the map error is an important outcome of this method, the crucial advantage is in the estimation consistency offered by the method. This consistency is a very crucial feature for planning purposes. A planner might be able to handle large errors as long as the filter indicates the estimates are unreliable (e.g., via it variance). But, if the filter returns a wrong estimate with low variance (i.e., it is confident that its estimate is correct, while it is not), then the planner most probably fail. 

\begin{figure}[ht!]
	\centering
	\subfigure{\includegraphics[width=.8\columnwidth]{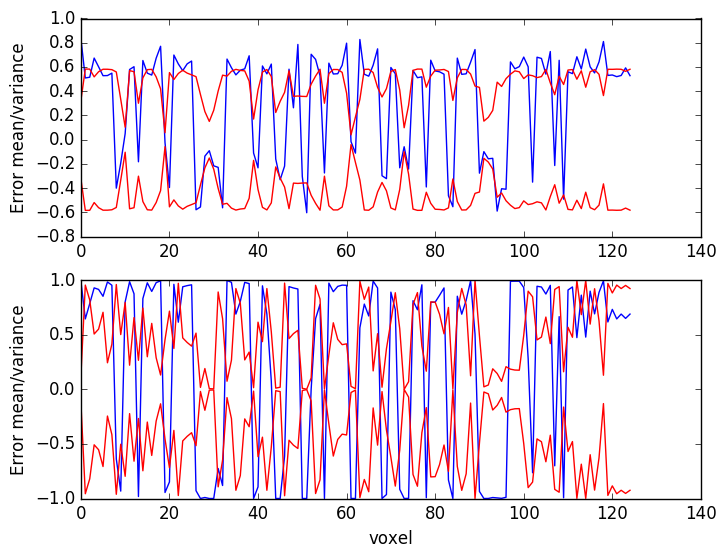}\label{fig:bad_error_consistency}}
	\subfigure{\includegraphics[width=.7\columnwidth]{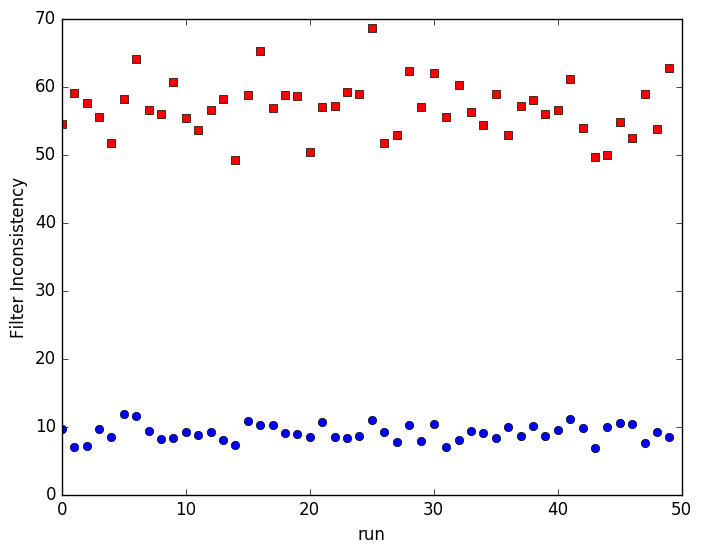}\label{fig:inconsistency}}
	\caption{(a) Mean (blue) and variance ($ 3\sigma $ confidence bound -- in red) of the mapping error over a set of voxels. Top axes correspond to the proposed method and the bottom axes correspond to the log-odds method. (b) Inconsistency measure of the proposed method (blue) and ISM (red) over 50 random maps.}
\end{figure}

Figure \ref{fig:bad_error_consistency} shows the error value (blue) along with the $ 2\sigma $ error bound (red) corresponding to the map in Fig. \ref{fig:example-map}. This graph is limited to a subset of voxels where the error is above a certain threshold. The top and bottom axes correspond to the proposed and ISM-based method, respectively. As it can be seen from top axes in Fig. \ref{fig:bad_error_consistency}, the $ 2\sigma $ bound in the proposed method grows and shrinks in a consistent manner with the error and behaves as a consistent confidence interval that can be used in planning. However, in bottom axes, where log-odds results are shown, there exist many voxels where the error is high (close to -1 or 1, i.e., free has been estimated as fully occupied or vice versa) \textit{at which the variance is very low}, which pose a significant challenge to the planner. 

To further study the filter consistency, we generate 50 environments and run the robot and mapping algorithm. For each run, we compute the inconsistency measure defined in Eq. \eqref{eq:inconsistencyMeasure}. The results are shown in Fig. \ref{fig:inconsistency}. Red squares show the value of $ I_c $ for ISM-based planning. Blue dots correspond to the proposed method. The gap shows a high improvement in estimation consistency, which is a desirable feature for planning methods.

\section{Conclusion} \label{sec:conclusion}
This paper proposes an alternative algorithm for occupancy grid mapping, by storing a richer data in the map. It extends traditional grid mapping in a few ways: first, relaxes the full-independence assumption and incorporates dependence between voxels in the measurement cone into the mapping scheme. Second, it relaxes the need for hand-engineering inverse sensor model and proposes the sensor cause model that can be derived from the forward sensor model. Third, it provides a consistent confidence values over occupancy estimation that can be reliably used in planning. The method runs online as measurements are received. Finally, it also enables mapping environments where voxels might be partially occupied. Results show that the mapping accuracy is often better than the ISM-based mapping, and more importantly the confidence values are much more reliable than traditional ISM-based mapping. 

The paper also discusses the joint design of grid-based mapping and planning methods, where the proposed mapping algorithm enables the planning method to predict the future evolution of the map under different candidate trajectories. Similar to POMDP-based literature on Simultaneous Localization and Plannign (SLAP) (e.g., \cite{Agha14-SLAP-Arxiv,Ali14-IJRR,Ali14-RolloutFIRM-ICRA}), we believe this work paves the way for future POMDP-based frameworks to solve Simultaneous Mapping and Planning (SMAP) in a more principled way.

\bibliographystyle{ieeeTran} 
\bibliography{AliAgha}

\begin{thebibliography}{10}
\providecommand{\url}[1]{#1}
\csname url@samestyle\endcsname
\providecommand{\newblock}{\relax}
\providecommand{\bibinfo}[2]{#2}
\providecommand{\BIBentrySTDinterwordspacing}{\spaceskip=0pt\relax}
\providecommand{\BIBentryALTinterwordstretchfactor}{4}
\providecommand{\BIBentryALTinterwordspacing}{\spaceskip=\fontdimen2\font plus
\BIBentryALTinterwordstretchfactor\fontdimen3\font minus
  \fontdimen4\font\relax}
\providecommand{\BIBforeignlanguage}[2]{{%
\expandafter\ifx\csname l@#1\endcsname\relax
\typeout{** WARNING: IEEEtran.bst: No hyphenation pattern has been}%
\typeout{** loaded for the language `#1'. Using the pattern for}%
\typeout{** the default language instead.}%
\else
\language=\csname l@#1\endcsname
\fi
#2}}
\providecommand{\BIBdecl}{\relax}
\BIBdecl

\bibitem{moravec1988sensor}
H.~P. Moravec, ``Sensor fusion in certainty grids for mobile robots,'' \emph{AI
  magazine}, vol.~9, no.~2, p.~61, 1988.

\bibitem{Elfes1989occupancy}
A.~Elfes, ``Occupancy grids: A probabilistic framework for robot perception and
  navigation,'' Ph.D. dissertation, Carnegie Mellon University, 1989.

\bibitem{Thrun2005}
S.~Thrun, W.~Burgard, and D.~Fox, \emph{Probabilistic Robotics}.\hskip 1em plus
  0.5em minus 0.4em\relax MIT Press, Cambridge, MA, 2005.

\bibitem{stachniss2009_book}
C.~Stachniss, \emph{Robotic mapping and exploration}.\hskip 1em plus 0.5em
  minus 0.4em\relax Springer, 2009, vol.~55.

\bibitem{thrun2002robotic}
S.~Thrun \emph{et~al.}, ``Robotic mapping: A survey,'' \emph{Exploring
  artificial intelligence in the new millennium}, vol.~1, pp. 1--35.

\bibitem{konolige2008outdoor}
K.~Konolige, M.~Agrawal, R.~C. Bolles, C.~Cowan, M.~Fischler, and B.~Gerkey,
  ``Outdoor mapping and navigation using stereo vision,'' in \emph{Experimental
  Robotics}.\hskip 1em plus 0.5em minus 0.4em\relax Springer, 2008, pp.
  179--190.

\bibitem{yamauchi1997frontier}
B.~Yamauchi, ``A frontier-based approach for autonomous exploration,'' in
  \emph{IEEE International Symposium on Computational Intelligence in Robotics
  and Automation}, 1997, pp. 146--151.

\bibitem{thrun1998learning}
S.~Thrun, ``Learning metric-topological maps for indoor mobile robot
  navigation,'' \emph{Artificial Intelligence}, vol.~99, no.~1.

\bibitem{wurm2010octomap}
K.~M. Wurm, A.~Hornung, M.~Bennewitz, C.~Stachniss, and W.~Burgard, ``Octomap:
  A probabilistic, flexible, and compact 3d map representation for robotic
  systems,'' in \emph{ICRA 2010 workshop on best practice in 3D perception and
  modeling for mobile manipulation}, vol.~2, 2010.

\bibitem{newcombe2011kinectfusion}
R.~A. Newcombe, S.~Izadi, O.~Hilliges, D.~Molyneaux, D.~Kim, A.~J. Davison,
  P.~Kohi, J.~Shotton, S.~Hodges, and A.~Fitzgibbon, ``Kinectfusion: Real-time
  dense surface mapping and tracking,'' in \emph{Mixed and augmented reality
  (ISMAR), 2011 10th IEEE international symposium on}.\hskip 1em plus 0.5em
  minus 0.4em\relax IEEE, 2011, pp. 127--136.

\bibitem{howard1996generating}
A.~Howard and L.~Kitchen, ``Generating sonar maps in highly specular
  environments,'' in \emph{In Proceedings of the Fourth International
  Conference on Control Automation Robotics and Vision}, 1996.

\bibitem{moravec1996TR}
H.~P. Moravec, ``Robot spatial perception by stereoscopic vision and 3d
  evidence grids,'' \emph{Technical Report CMU-RI-TR-96-34, Carnegie Mellon
  University}, 1996.

\bibitem{pagac1996evidential}
D.~Pagac, E.~M. Nebot, and H.~Durrant-Whyte, ``An evidential approach to
  probabilistic map-building,'' in \emph{Reasoning with Uncertainty in
  Robotics}.\hskip 1em plus 0.5em minus 0.4em\relax Springer, 1996, pp.
  164--170.

\bibitem{konolige1997improved}
K.~Konolige, ``Improved occupancy grids for map building,'' \emph{Autonomous
  Robots}, vol.~4, no.~4, pp. 351--367, 1997.

\bibitem{Paskin05}
P.~M and T.~S, ``Robotic mapping with polygonal random fields,'' in
  \emph{Conference on Uncertainty in Artificial Intelligence}, 2005, p.
  450–458.

\bibitem{veeck2004learning}
M.~Veeck and W.~Burgard, ``Learning polyline maps from range scan data acquired
  with mobile robots,'' in \emph{IEEE/RSJ International Conference on
  Intelligent Robots and Systems}, vol.~2, 2004, pp. 1065--1070.

\bibitem{thrun2003learning}
S.~Thrun, ``Learning occupancy grid maps with forward sensor models,''
  \emph{Autonomous robots}, vol.~15, no.~2, pp. 111--127, 2003.

\bibitem{kay1993fundamentals}
S.~M. Kay, ``Fundamentals of statistical signal processing, volume i:
  estimation theory,'' 1993.

\bibitem{Lavalle01-RRT}
S.~Lavalle and J.~Kuffner, ``Randomized kinodynamic planning,''
  \emph{International Journal of Robotics Research}, vol.~20, no. 378-400,
  2001.

\bibitem{Agha14-SLAP-Arxiv}
\BIBentryALTinterwordspacing
A.~Agha{-}mohammadi, S.~Agarwal, S.~Chakravorty, and N.~M. Amato, ``{SLAP}:
  Simultaneous localization and planning for physical mobile robots via
  enabling dynamic replanning in belief space,'' \emph{CoRR}, vol.
  abs/1510.07380, 2015. [Online]. Available:
  \url{http://arxiv.org/abs/1510.07380}
\BIBentrySTDinterwordspacing

\bibitem{Ali14-IJRR}
A.~{Agha-mohammadi}, S.~Chakravorty, and N.~Amato, ``{FIRM}: Sampling-based
  feedback motion planning under motion uncertainty and imperfect
  measurements,'' \emph{International Journal of Robotics Research (IJRR)},
  vol.~33, no.~2, pp. 268--304, 2014.

\bibitem{Ali14-RolloutFIRM-ICRA}
A.~{Agha-mohammadi}, S.~Agarwal, A.~Mahadevan, S.~Chakravorty, D.~Tomkins,
  J.~Denny, and N.~Amato, ``Robust online belief space planning in changing
  environments: Application to physical mobile robots,'' in \emph{IEEE Int.
  Conf. Robot. Autom. (ICRA)}, Hong Kong, China, 2014.

\end{thebibliography}

\vspace{-10pt}
\begin{appendices}
	
	\section{Proof of Lemma \ref{lem:cause-meas}} \label{app:proofLemma}
	\vspace{-20pt}
	\begin{align}
	\nonumber
	&p(m^i|c_k,z_{0:k},xv_{0:k})
	\\
	\nonumber
	&=
	\frac{p(z_k|m^i,c_{k},z_{0:k-1},xv_{0:k})p(m^i|c_{k},z_{0:k-1},xv_{0:k})}
	{p(z_k|c_{k},z_{0:k-1},xv_{0:k})}
	\\
	\nonumber
	&=
	\frac{p(z_k|c_{k},z_{0:k-1},xv_{0:k})p(m^i|c_{k},z_{0:k-1},xv_{0:k})}
	{p(z_k|c_{k},z_{0:k-1},xv_{0:k})}
	\\
	\nonumber &=
	p(m^i|c_{k},z_{0:k-1},xv_{0:k})
	\end{align}
	

\end{appendices}

\end{document}